\documentclass[runningheads,a4paper]{llncs}

\usepackage{amssymb}
\usepackage{graphicx}
\usepackage{wrapfig}
\usepackage{url}
\urldef{\mailsa}\path|ahalyaprabhakar2013@u.northwestern.edu;  stacymav@u.northwestern.edu;
jschultz@northwestern.edu; t-murphey@northwestern.edu|
\urldef{\mailsb}\path|anna.kramer, leonie.kunz, christine.reiss, nicole.sator,|
\urldef{\mailsc}\path|erika.siebert-cole, peter.strasser, lncs}@springer.com|    
\newcommand{\keywords}[1]{\par\addvspace\baselineskip
\noindent\keywordname\enspace\ignorespaces#1}
\usepackage[numbers]{natbib}

\begin{document}

\mainmatter  

\title{Autonomous Visual Rendering using Physical Motion}

\titlerunning{Autonomous Visual Rendering using Physical Motion}

%
%
\author{Ahalya Prabhakar%
\and Anastasia Mavrommati\and Jarvis Schultz\and Todd D. Murphey}
\authorrunning{Autonomous Visual Rendering using Physical Motion}

\institute{Department of Mechanical Engineering, Northwestern University,\\
	 2145 	Sheridan Road, Evanston, IL 60208, USA\\
\mailsa}

%
%

\toctitle{Autonomous Visual Rendering using Physical Motion}
\tocauthor{Authors' Instructions}
\maketitle

\begin{abstract}
\emph{This paper addresses the problem of enabling a robot to represent and recreate visual information through physical motion, focusing on drawing using pens, brushes, or other tools. This work uses ergodicity as a control objective that translates planar visual input to physical motion without preprocessing (e.g., image processing, motion primitives). 
 We achieve comparable results to existing drawing methods, while reducing the algorithmic complexity of the software. We demonstrate that optimal ergodic control algorithms with different time-horizon characteristics (infinitesimal, finite, and receding horizon) can generate qualitatively and stylistically different motions that render a wide range of visual information (e.g., letters, portraits, landscapes). In addition, we show that ergodic control enables the same software design to apply to multiple robotic systems by incorporating their particular dynamics, thereby reducing the dependence on task-specific robots. Finally, we demonstrate physical drawings with the Baxter robot. } 
\end{abstract}

 \section{Introduction}
\label{Introduction}

An increasing amount of research is focused on using control theory as a generator of artistic expressions for robotics applications \cite{laviers2014controls}.
There is a large interest in enabling robots to create art, such as drawing \cite{jean2012artist}, dancing \cite{laviers2011automatic}, or writing. However, the computational tools available in the standard software repertoire are generally insufficient for enabling these tasks in a natural and interpretable manner. This paper focuses on enabling robots to draw and write by translating raw visual input into physical actions. 

Drawing is a task that does not lend itself to analysis in terms of trajectory error.  Being at a particular state at a particular time does not improve a drawing, and failing to do so does not make it worse.  Instead, drawing is a process where the success or failure is determined after the entire time history of motion has been synthesized into a final product.  How should ``error'' be defined for purpose of quantitative engineering decisions and software automation?  Similarly, motion primitives can be an important foundation for tasks such as drawing (e.g., hatch marks to represent shading), but where should these primitives come from and what should be done if a robot cannot physically execute them? Questions such as these often lead to robots and their software being co-designed with the task in mind, leading to task-specific software enabling a task-specific robot to complete the task.  How can we enable drawing-like tasks in robots as they \emph{are} rather than as we would \emph{like them to be}?  And how can we do so in a manner that minimizes tuning (e.g., in the case of drawing the same parameters can be used for both faces and landscapes) while also minimizing software complexity? \emph{In this paper we find that the use of ergodic metrics---and the resulting ergodic control---reduces the dependence on task-specific robots (e.g., robots mechanically designed with drawing in mind), reduces the algorithmic complexity of the software that enables the task (e.g., the number of independent processes involved in drawing decreases), and enables the same software solution to apply to multiple robotic instantiations.}

Moreover, this paper touches on a fundamental issue for many modern robotic systems---the need to communicate through motion.  Symbolic representations of information are the currency of communication, physically transmitted through whatever communication channels are available (electrical signals, light, body language, written language and related symbolic artifacts such as drawings).  The internal representation of a symbol must both be \emph{perceivable} given a sensor suite (voltage readings, cameras, tactile sensors) and \emph{actionable} given an actuator suite (signal generators, motors). \emph{Insofar as all systems can execute ergodic control, we hypothesize that ergodic metrics provide a nearly-universal, actionable measure of spatially-defined symbolic information.} Specifically, in this paper we see that both letters (represented in a font) and photographs can be rendered by a robotic system working within its own particular physical capabilities. For instance, a hand-writing-like rendering of the letter “N” (seen later in Figure \ref{fig:N_single_traj}) is seen to be a consequence of putting a premium on efficiency for a first-order dynamical system rendering the letter.  Moreover, in the context of drawing photographs (of people and landscapes), we see a) that other drawing algorithms implicitly optimize (or at least improve) ergodicity, and b) using ergodic control, multiple dynamical systems approach rendering in dramatically different manners with similar levels of success. 

We begin by introducing ergodicity in Section \ref{sec:Ergodicity}, including a discussion of its characteristics. Section \ref{sec:Control Methods} includes an overview and comparison of the ergodic control methods used in this paper. We present some examples in Section \ref{sec:Examples}, including comparisons of the results of the different ergodic control schemes introduced in the previous section, comparisons with existing robot drawing methods, and experiments using the Baxter robot. 

\section{Methods}
\label{Methods}

\subsection{Ergodicity}
\label{sec:Ergodicity}

\begin{wrapfigure}{r}{0.5\textwidth}
    \centering
    \includegraphics[width = .3 \textwidth]{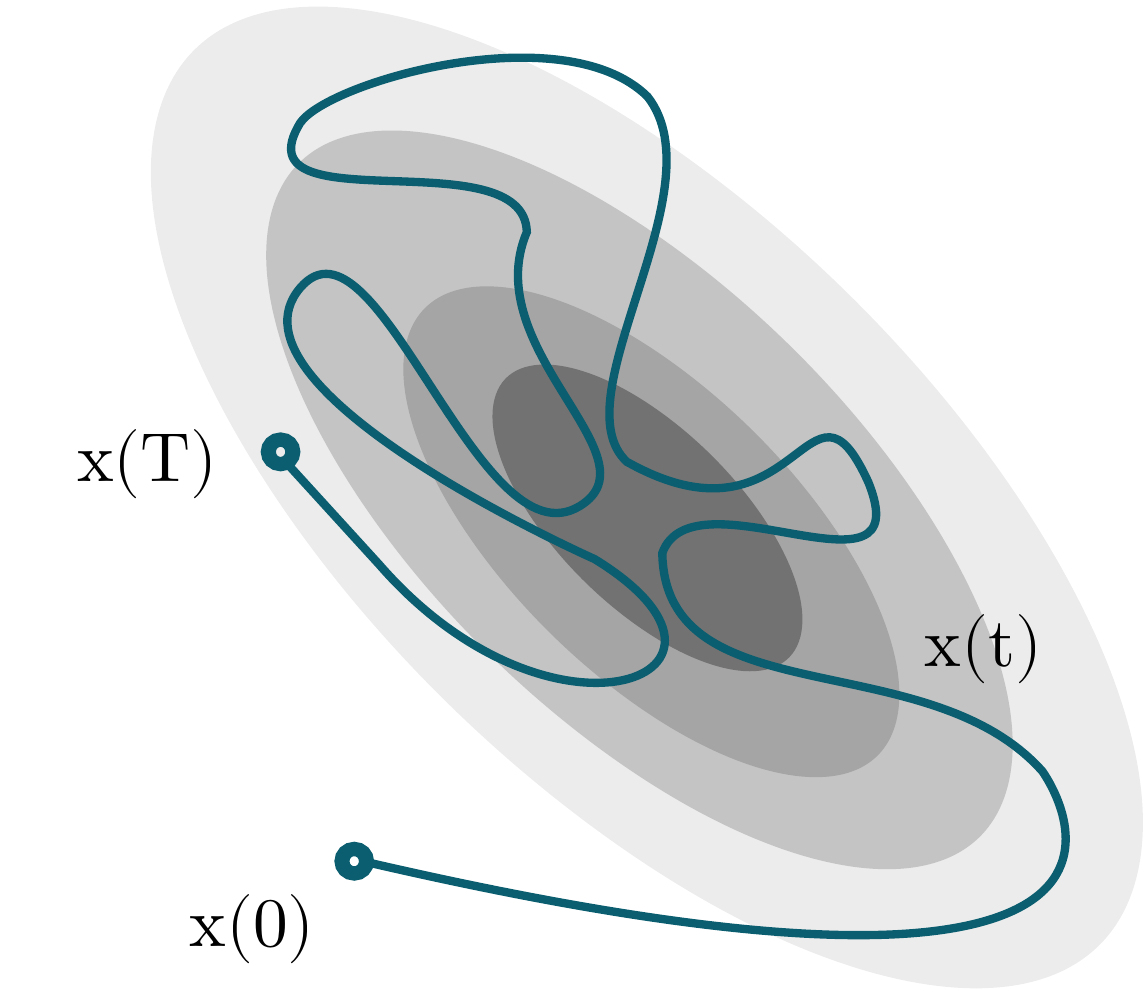}
    \caption{An illustration of ergodic trajectories. Ergodic trajectories spend time in the workspace proportional to the spatial distribution. }\label{fig:ergvsnonerg}
\end{wrapfigure}

Ergodicity compares the spatial statistics of a trajectory to the spatial statistics of a desired spatial distribution. A trajectory is \textit{ergodic} with respect to a spatial distribution if the time spent in a region is proportional to the density of the spatial distribution. In previous work, the ergodic metrics have been employed for other applications including search \cite{miller2016ergodic,MillerACC2013} or coverage \cite{shell2006ergodic}. When used for search-based applications, ergodicity encodes the idea that the higher the information density of a region in the distribution, the more time spent in that region, shown in Figure \ref{fig:ergvsnonerg}.
  
The spatial distributions used in this paper represent the spatial distribution of the symbol or image being recreated through motion, introduced in \cite{sahai2015chaotic}. The more intense the color in the image, the higher the value of the spatial distribution. Thus, ergodicity encodes the idea that the trajectory represents the path of a tool (e.g., marker, paintbrush, etc.), where the longer the tool spends drawing in the region the greater the intensity of the color in that region. 

To evaluate the ergodicity, we define the ergodic metric to be the \textit{distance from ergodicity $\varepsilon$} of the time-averaged trajectory from the spatial distribution $\phi(x)$. The ergodicity of the trajectory is computed as the sum of the weighted squared distance between the Fourier coefficients of the spatial distribution $\phi_{k}$ and the distribution representing the time-averaged trajectory $c_{k}$, defined below: 
\small
\begin{equation} \label{eq:ergodicityMetric}
\varepsilon =\sum_{k_{1}=0}^{K}\dots\sum_{k_{n}=0}^{K} \Lambda_{k}|c_{k}-\phi_{k}|^{2} ,
\end{equation} 
\normalsize  
 where $K$ is the number of coefficients calculated along each of the $n$ dimensions, and $k$ is a multi-index $k=(k_{1},...,k_{n})$.  The coefficients $\Lambda_{k}$ weight the lower frequency information higher and are defined as $\Lambda_{k} = \frac{1}{(1+||k||^{2})^{s}}$, where $s = \frac{n+1}{2}$.

The Fourier basis functions are determined as below:
\small
\begin{equation}\label{eq:fourierbasisfxns}
F_{k}(x) = \frac{1}{h_{k}}\prod_{i=1}^{n}\cos \left(\frac{k_{i}\pi}{L_{i}}x_{i}\right),
\end{equation} 
\normalsize
where $h_{k}$ is a normalizing factor as defined in \citep{MathewMezic2011}.
The spatial Fourier coefficients are computed from the inner product 
\small 
\begin{equation}
\phi_{k} = \int_{X}\phi(x)F_{k}(x)dx,
\end{equation}
\normalsize
 and the Fourier coefficients of the trajectory $x(\cdot)$ are evaluated as  
\begin{equation}
\small
c_{k} = \frac{1}{T}\int_{0}^{T}F_{k}(x(t))dt,
\end{equation}
\normalsize
where $T$ is the final time of the trajectory \cite{MathewMezic2011}. 
\subsection{Ergodic Control Algorithms}
\label{sec:Control Methods}

To demonstrate the different styles of resulting motions obtained from different methods, we compare the results of three ergodic control algorithms. All three algorithms generate trajectories that 
reduce the ergodic cost in (\ref{eq:ergodicityMetric}), but each exhibits different time-horizon characteristics.  

The algorithm with an infinitesimally small time horizon is a closed-form ergodic control (CFEC) method derived in \citep{MathewMezic2011}. At each time step, the feedback control is calculated as the closed-form solution to the optimal control problem with ergodic cost in the limit as the receding horizon goes to zero. The optimal solution is obtained by minimizing a Hamiltonian \cite{kirk2012optimal}. Due to its receding-horizon origins, the resulting control trajectories are piecewise continuous. The method can only be applied to linear first-order and second-order dynamics, with constant speed and forcing respectively. Thus, CFEC is an ergodic control algorithm that optimizes ergodicity along an infinitesimal time horizon at every time step in order to calculate the next control action.

The algorithm with a non-zero receding time horizon is Ergodic iterative-Sequential Action Control (E-iSAC), based on Sequential Action Control (SAC) \cite{Ansari2016TR0, Ansari2016SAC}. At each time step, E-iSAC uses hybrid control theory to calculate the control action that optimally improves ergodicity over a non-zero receding time horizon. Like CFEC, resulting controls are piecewise continuous. The method can generate ergodic trajectories for both linear and nonlinear dynamics, with saturated controls.

Finally, the method with a non-receding, finite time horizon is the ergodic Projection-based Trajectory Optimization (PTO) method derived in \cite{MillerACC2013, miller2016ergodic}.  Unlike the previous two approaches, it is an infinite-dimensional gradient-descent algorithm that outputs continuous control trajectories. Like E-iSAC, it can take into account the linear/nonlinear dynamics of the robotic system, but it calculates the control trajectory over the entire time duration that most efficiently minimizes the ergodic metric rather than simply the next time step. It also has a weight on control in its objective function that balances the ergodic metric, to achieve a dynamically-efficient trajectory that minimizes the ergodic metric.

\begin{table}
\centering
  \begin{tabular}{ | c || c | c | c |}
    \hline
    \textbf{ Features }  & \textbf{ CFEC } & \textbf{ E-iSAC } & \textbf{ PTO } \\  \hline
   Finite Time Horizon   &           &           & $\bullet$ \\ \hline 
   Closed Loop Control   & $\bullet$ & $\bullet$ &           \\ \hline
   Nonlinear Dynamics    &           & $\bullet$ & $\bullet$ \\ \hline
   Control Saturation    & $\bullet$ & $\bullet$ &           \\ \hline
   Receding Horizon      &           & $\bullet$ &           \\ \hline
   Continuous Control    &           &           & $\bullet$ \\ \hline
   Weight on Control     &           & $\bullet$ & $\bullet$ \\ \hline   
   Efficient Computation & $\bullet$ & $\bullet$ &           \\ \hline 
  \end{tabular}

  \caption{Comparison of features for the different ergodic control methods used in the examples.}\label{tab: Method Feature Comparison}
\end{table}  

Both CFEC and E-iSAC are efficient to compute, whereas PTO is computationally expensive as it requires numerical integration of several complex differential equations for the entire finite time horizon during each iteration of the algorithm. Next, we investigate the application of these techniques to examples including both letters and photographs. 
\section{Examples}
\label{sec:Examples}
\subsection{Writing Symbols}
\label{sec:N_example}

In the first example, we investigate how a robot can use ergodic control  to recreate a structured visual input, such as a letter, presented as an image. In addition, because the input image is  not merely  of artistic interest but also  corresponds to a recognizable symbol (in this case, the letter ``N''), we show how  a robot can render meaningful visual cues, without the prior knowledge of a dictionary or library of symbols. To do this, we represented the image of the letter as a spatial distribution as described in Section \ref{sec:Ergodicity}. We then determined the trajectories for the three different methods (CFEC, E-iSAC, and PTO) described in Section \ref{sec:Control Methods} for systems with first order dynamics and second order dynamics. We ran all the simulations for 60 seconds total with the same number of Fourier coefficients to represent the image spatially.

\begin{figure}[h!]
    \centering
    \includegraphics[width = .65\textwidth]{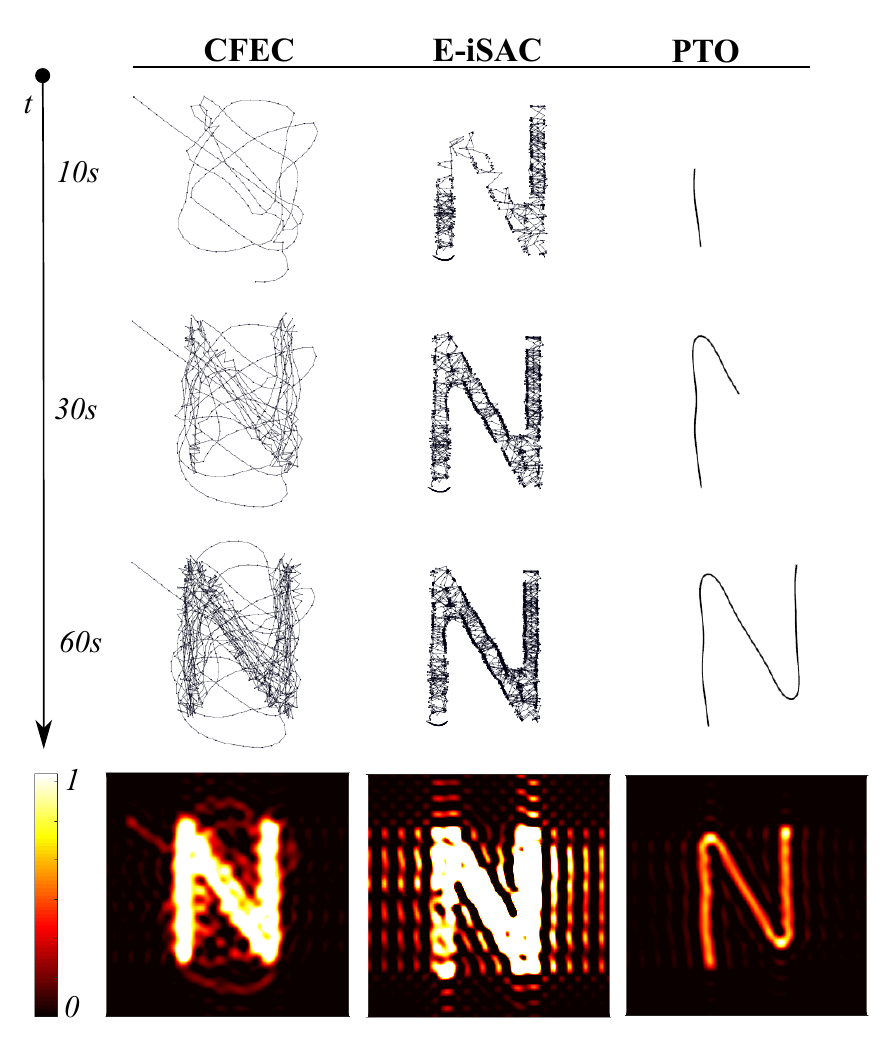}
    \caption{ Trajectories generated by the three different methods with single integrator dynamics for the letter ``N'' at 10 seconds, 30 seconds and 60 seconds with the spatial reconstructions of the time-averaged trajectories generated by each of the methods at the final time. The different methods lead to stylistic differences with an abstract representation from the CFEC method to a natural, human-penmanship motion from the PTO method.}\label{fig:N_single_traj}
\end{figure} 

Figure \ref{fig:N_single_traj} shows the resulting ergodic motions for each control algorithm with the drawing dynamics represented as a single integrator. From Figure \ref{fig:N_single_traj}, we can see that while all three methods produce results that are recognizable as the letter ``N'', the trajectories generated to achieve this objective are drastically different.
The velocity-control characteristic of the single integrator system leads to the sharp, choppy turns evident in both discrete-time CFEC and E-iSAC methods. The infinitesimally small time horizon of the CFEC method, in contrast to the non-zero receding horizon of the E-iSAC method, results in the large, aggressive motions of the CFEC result compared to the E-iSAC result. Finally, the weight on the control cost and the continuous-time characteristics of the PTO method lead to a rendering that most closely resembles typical human penmanship.

\begin{figure}[h!]
    \centering
    \includegraphics[width = .65\textwidth]{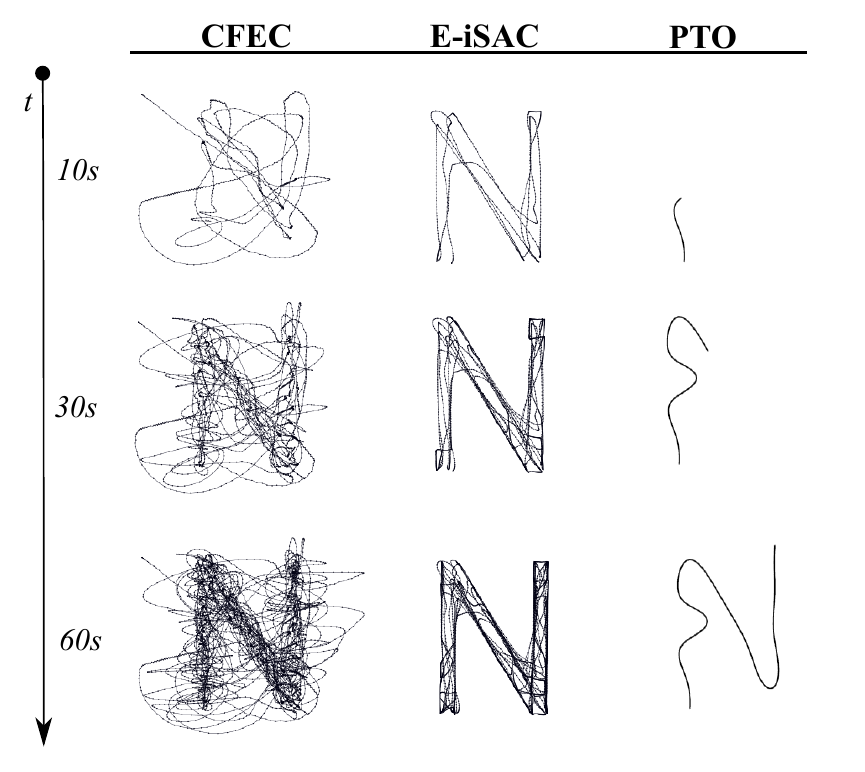}
    \caption{The trajectories generated by the three different methods with double integrator dynamics for the letter ``N'' at 10 seconds, 30 seconds and 60 seconds. The double-order dynamics lead to much smoother motion from the discrete methods (CFEC and E-iSAC) and more ergodic results from the E-iSAC methods due to its receding-horizon characteristics.   }\label{fig:N_double_traj}
\end{figure} 

For the double integrator system shown in Figure \ref{fig:N_double_traj}, the controls are accelerations rather than the velocities of the system. Because of this, the trajectories produced by the discrete controls  for the CFEC and E-iSAC method are much smoother, without the sharp turns seen in Figure \ref{fig:N_single_traj}.
Even though the CFEC result is smoother, its trajectory is more abstract and messy than the single integrator trajectory. The receding horizon E-iSAC produces much better results for systems with complicated or free dynamics (e.g., drift), including the double integrator system. The trajectory produced executes an ``N'' motion reminiscent of human penmanship and continues to draw similarly smooth motions over the time horizon. While PTO leads to a similarly smooth result compared to the single integrator system, it leads to a result that less resembles human penmanship.  

From both examples, we can see how ergodicity can be used as a representation of symbolic spatial information (i.e., the letter ``N'') and ergodic control algorithms can be used to determine the actions needed to sufficiently render the information while incorporating the physical capabilities of the robot.  
  
\begin{figure}[h!]
    \centering
    \includegraphics[width = \textwidth]{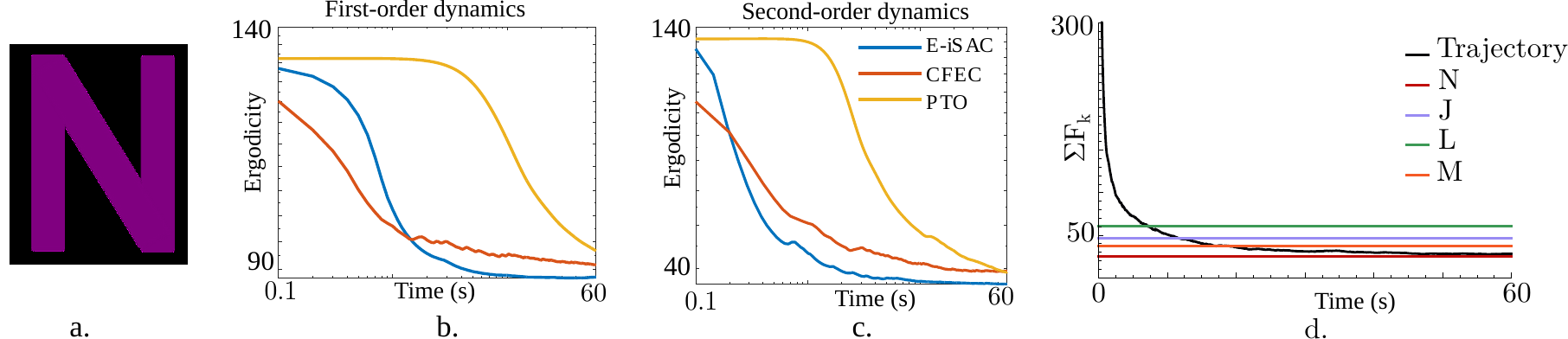}
    \caption{a)-c) Time evolution of the ergodic metric, or the normalized difference of the squared sum of the trajectory Fourier  coefficients and the spatial Fourier  coefficients, for the three different methods with first-order and second-order dynamics for the letter ``N'' on logarithmic scale. Note that because ergodicity can only be calculated over a state trajectory of finite time duration (see Eq. \ref{eq:ergodicityMetric}), we start measuring the ergodicity values once 0.1 seconds of simulation have passed; hence the three approaches start at different ergodic metrics. All three methods result in similarly ergodic trajectories by the end of the time horizon, with differences due to their time-horizon characteristics. d) Sum of the Fourier coefficients for the trajectory over time compared to the spatial Fourier coefficients of different letters (N, J, L, and M). The trajectory coefficients converge to the spatial coefficients of the letter ``N'' that is being drawn, quantitatively representing the process of discriminating the specific symbol being drawn over time. }\label{fig:N_ergodicities_letter}
\end{figure}

Figure \ref{fig:N_ergodicities_letter}a-\ref{fig:N_ergodicities_letter}c shows the ergodic metric, or the difference between the sum of the trajectory Fourier  coefficients and spatial Fourier coefficients, for the different methods over time. We can see that for both dynamic systems, all three methods produce trajectories that are similarly ergodic with respect to the letter by the end of the time horizon. Compared to E-iSAC, CFEC converges more slowly when more complex dynamics (i.e., double order dynamics) are introduced. PTO exhibits a lower rate of cost reduction because it performs finite-time horizon optimization. Note that E-iSAC always reaches the lowest ergodic cost value by the end of the 60 second simulation.

Figure \ref{fig:N_ergodicities_letter}d compares the sum of the absolute value of the Fourier coefficients trajectory generated by the CFEC method for the single integrator to the spatial Fourier  coefficients of different letters. Initially, the difference between the trajectory coefficients and spatial Fourier  coefficients is large, representing the ambiguity of the symbol being drawn. As the symbol becomes more clear, the Fourier coefficients of the trajectory converge to the coefficients of ``N'', representing the discrimination of the letter being drawn from the other letters. 
Moreover, letters that are more visually similar to the letter ``N'' have Fourier coefficients that are quantitatively closer to the the spatial coefficients of ``N'' and thus take longer to distinguish which symbol is being drawn.
The ergodic control metric allows for representation of a symbol from visual cues and discrimination of that symbol from others without requiring prior knowledge of the symbols. 
   
\subsection{Drawing Images}
\label{sec:Lincoln_example}
In the second example, we consider drawing a picture from a photograph as opposed to drawing a symbol. Previously, we were concerned with the representation of the structured symbol. In this example, we move to representing a more abstract image for purely artistic expression. Here, we are drawing a portrait of Abraham Lincoln\footnote{The image was obtained from \url{https://commons.wikimedia.org/wiki/File:Abraham_Lincoln_head_on_shoulders_photo_portrait.jpg}.} with all three methods for single-order and double-order dynamics. We also render the portrait with a lightly damped spring system. The simulations are performed for the same 60-second time horizon and number of coefficients as the previous example.

\begin{figure}[h]
    \centering
    \includegraphics[width =.9 \textwidth]{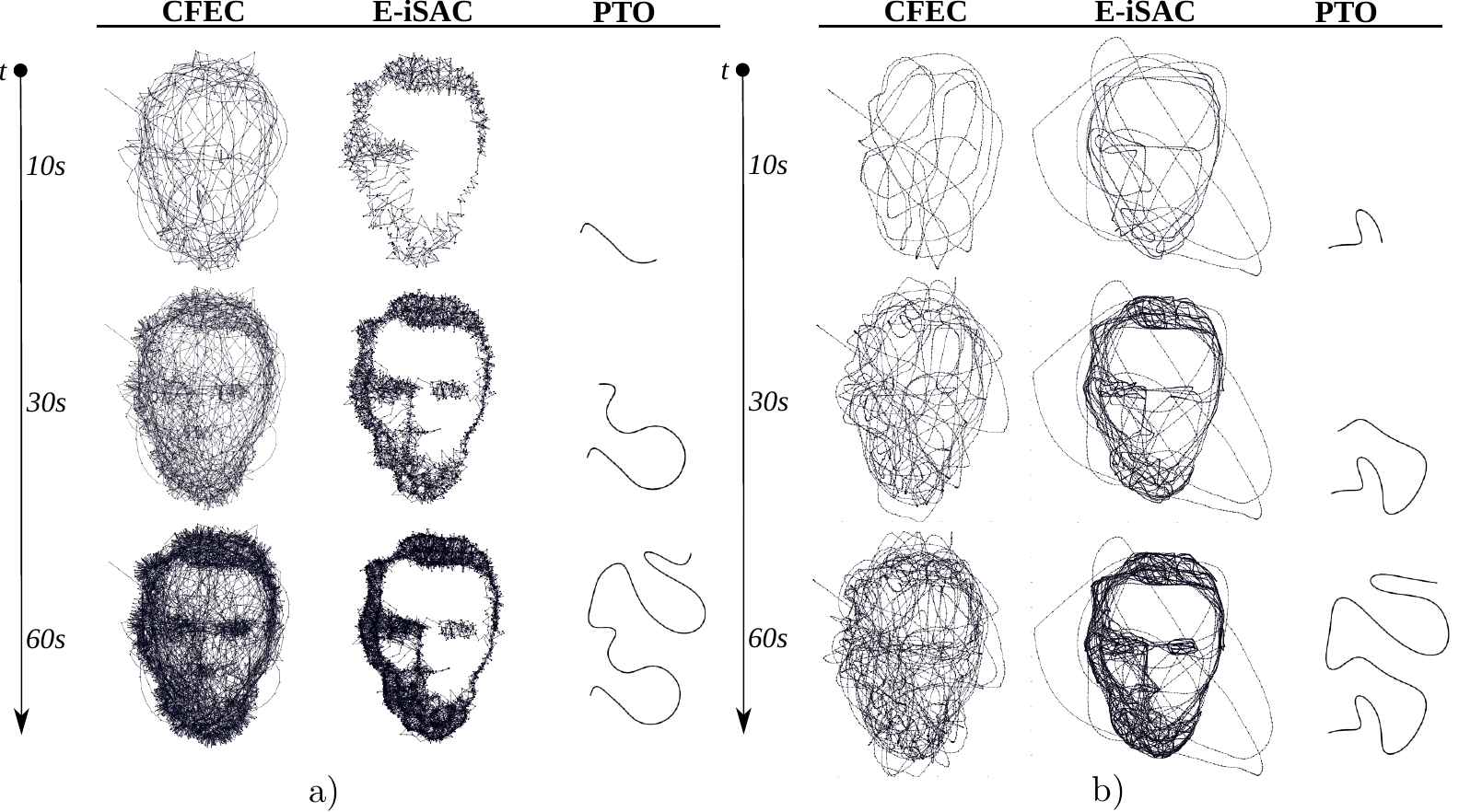}
    \caption{Trajectories generated by the three different methods for drawing the portrait of Abraham Lincoln at 10 seconds, 30 seconds and 60 seconds. a) Single Integrator Dynamics: CFEC and E-iSAC result in distinguishable results with stylistic differences, whereas PTO is inadequate for this purpose. b) Double Integrator Dynamics: E-iSAC results in a smooth, clear portrait of Abraham Lincoln with a trajectory that naturally draws the face as a person would sketch one, without any preprogramming or library of motion primitives. }\label{fig:linc_traj}
\end{figure} 

Figure \ref{fig:linc_traj}a compares the trajectories resulting from the different ergodic control methods for the single-integrator system. The weight on control and continuous-time characteristics of the PTO method that were desirable for the structured symbol example are disadvantageous in this case. While it reduces the ergodic cost, its susceptibility to local minima and its weight on energy lead to a far less ergodic result compared to the other methods. 

Instead, the discrete nature of the other two methods produce trajectories that are more clearly portraits of Lincoln and are more ergodic with respect to the photo. The trajectory produced by the CFEC method initially covers much of the area of the face, but returns to the regions of high interest such that the final image produced matches the original image closely in shading. The E-iSAC method produces a trajectory that is much cleaner and does not cover the regions that are not shaded in (i.e., the forehead, cheeks). The velocity control of the single-order system leads to a disjointed trajectory similar to the results from Fig. \ref{fig:N_single_traj}.

Figure \ref{fig:linc_traj}b compares the resulting trajectories for the double integrator system. As discussed previously, the PTO method produces a trajectory that is far less ergodic with respect to the photo. The control on acceleration significantly impacts the stylistic rendering of the CFEC rendering. The E-iSAC method produced better results than the other methods for the double integrator system, due to its longer time horizon. The resulting trajectory is smoother and more natural compared to the single integrator results. Interestingly, this method creates a trajectory that naturally draws the contours of the face--- the oval shape and the lines for the brows and nose--- before filling in the details, similar to the way that humans sketch a face \cite{loomis2011drawing}. 

\begin{figure}[h!]
    \centering
    \includegraphics[width =  \textwidth]{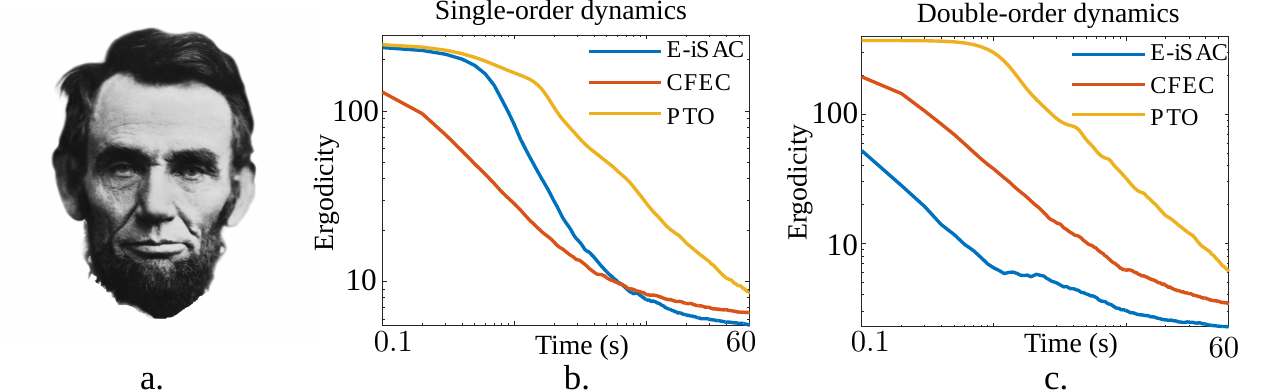}
    \caption{Time evolution of the ergodic metric, or the normalized difference of the squared sum of the trajectory Fourier  coefficients and the spatial Fourier  coefficients for the three different methods with single-order and double-order dynamics for the Lincoln image on logarithmic scale. E-iSAC results in a more optimally ergodic trajectory for both systems. PTO performs poorly for both systems and CFEC performs significantly worse with the double-order dynamics.  }\label{fig:linc_ergodicities}
\end{figure} 

Fig. \ref{fig:linc_ergodicities} compares the results of ergodicity with respect to time for the different methods. Similar to the symbolic example in Section \ref{sec:N_example}, the E-iSAC trajectory creates a more ergodic image by the end for both cases. While the CFEC method results in a less  ergodic trajectory for both systems, the ergodic cost for the single-order dynamics decreases faster for much of the time horizon. The CFEC method performs significantly worse for the double-order dynamics system and has a higher ergodic cost than the E-iSAC method throughout the entire time horizon. The inadequacy of the PTO method for this image is demonstrated here, having significantly higher ergodic costs at the final time for both systems.

\begin{figure}[h!]
    \centering
    \includegraphics[width =  \textwidth]{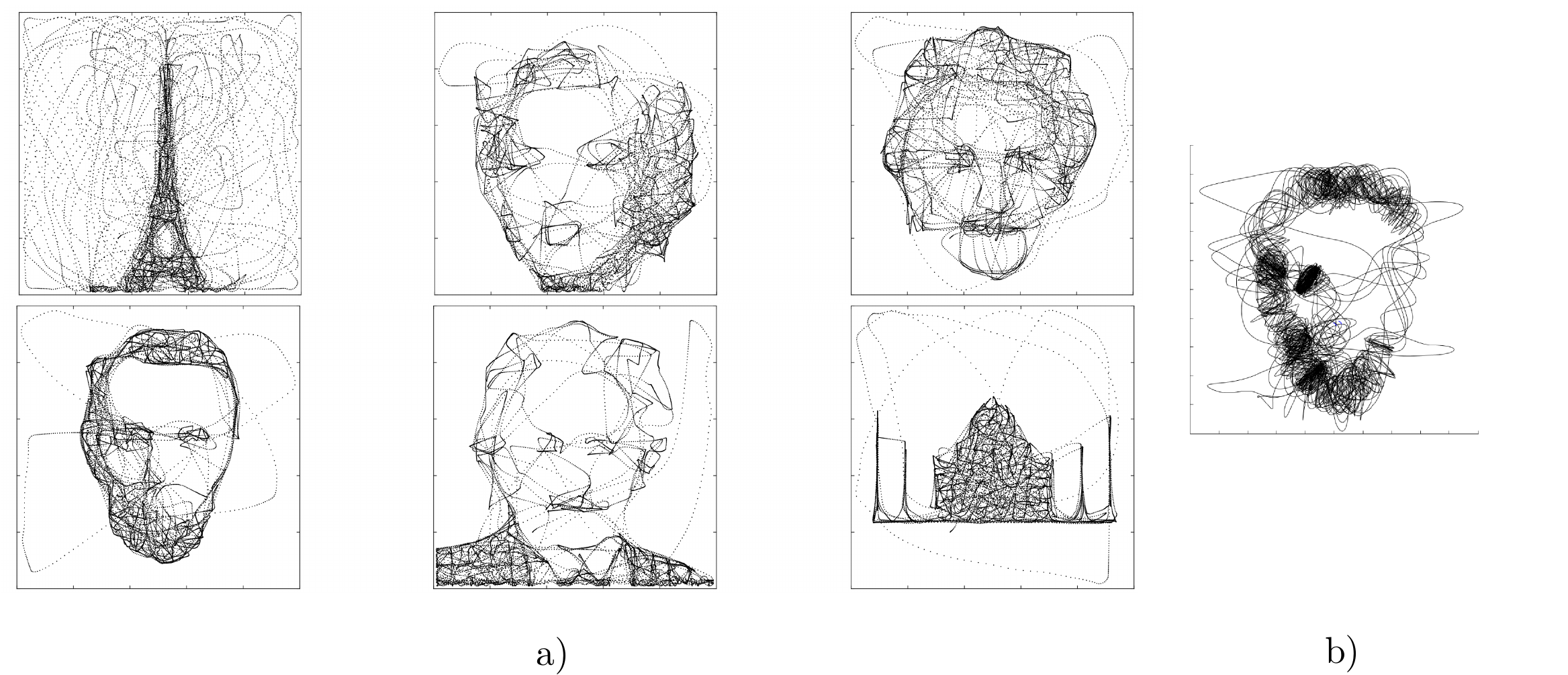}
    \caption{a) Renderings of different images (Eiffel tower, Marilyn Monroe, Einstein's face, Lincoln's face, Einstein with suit, and Taj Mahal) from the double-order dynamical system using the E-iSAC method with the same set of parameters. E-iSAC is able to successfully render different images with different content (faces and monuments) with the identical parameters. b) Trajectory generated by the E-iSAC method for the  Lincoln portrait image with damped spring dynamics. E-iSAC is able to produce a trajectory that reproduces the image while satisfying the dynamics of a system. }\label{fig:imgcompspring}
\end{figure} 

Figure \ref{fig:imgcompspring}a compares the renderings of six different images with different content--- portraits and monuments using the E-iSAC method with double-order dynamics. We show that E-iSAC successfully renders different images using a single set of parameters.

Finally, we demonstrate the ability of the E-iSAC method to take into account more complicated dynamics of the system. In Figure \ref{fig:imgcompspring}b, we simulate a system where the drawing mass is connected via a lightly damped spring to the controlled mass. The resulting Lincoln trajectory is harder to distinguish than the renderings from the single-order and double-order dynamical systems. However, the system is successfully able to optimize with respect to the dynamics and draw the main features of the face---the hair, the beard, and the eyes---within the time horizon, reducing the ergodic cost by $31.5\%$ in 60 seconds.

\subsection{Comparisons with Existing Robot Drawing Techniques}

Existing drawing robots employ a multistage process, using preprocessing (e.g., edge detection and/or other image processing techniques) and postprocessing (e.g., motion primitives, shade rendering, path planning) to render the image \cite{calinon2005humanoid, deussen2012feedback, jean2012artist,lu2009preliminary, tresset2013portrait}. To accomplish this, most robots and their software are co-designed specifically with drawing in mind, with most specializing in recreating scenes of specific structures, such as portraits of human faces. Similar multi-stage methods are commonly used for robot writing. They typically use image preprocessing, segmentation, waypoints, or a library of motion primitives to plan the trajectory and execute the trajectory using different motion control and trajectory tracking methods \citep{li2013human,Perez2015,Syamlan2015,yussof2005algorithm}.

\begin{figure}
    \centering
    \includegraphics[width =\textwidth]{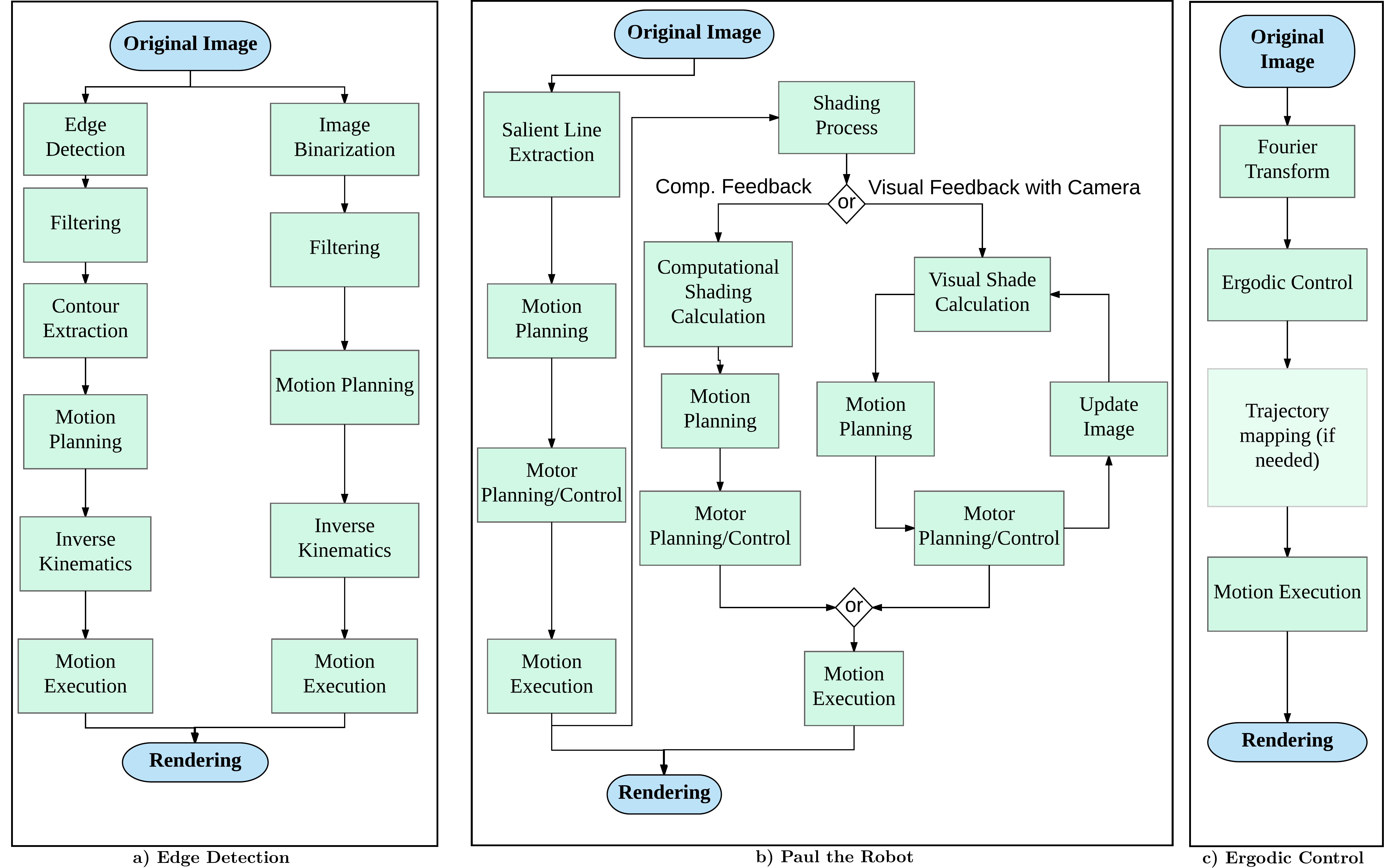}
    \caption{Comparison of Drawing Methods. a) Edge detection and Binarization method used in \cite{calinon2005humanoid} b) Method used by Paul the Drawing Robot from \cite{tresset2013portrait} and c) Ergodic Control. Ergodic Control is able to achieve comparable results with an algorithm requiring fewer independent processes. }\label{fig:Method_chart}
\end{figure} 

Recently, some approaches using motion-driven machine learning have been used to enable robots to learn and mimic human motion \citep{liang2015emg,Syamlan2015,yin2014learning}. These methods can be difficult and computationally costly and efforts to make them tractable (i.e., predefined dictionary of symbols) can be limiting in scope \cite{li2013human,yussof2005algorithm}.  Furthermore, they do not consider the robot's physical capabilities and can thus generate motions that are difficult for the robot to execute. 

\begin{figure}[h]
    \centering
    \includegraphics[width = \textwidth]{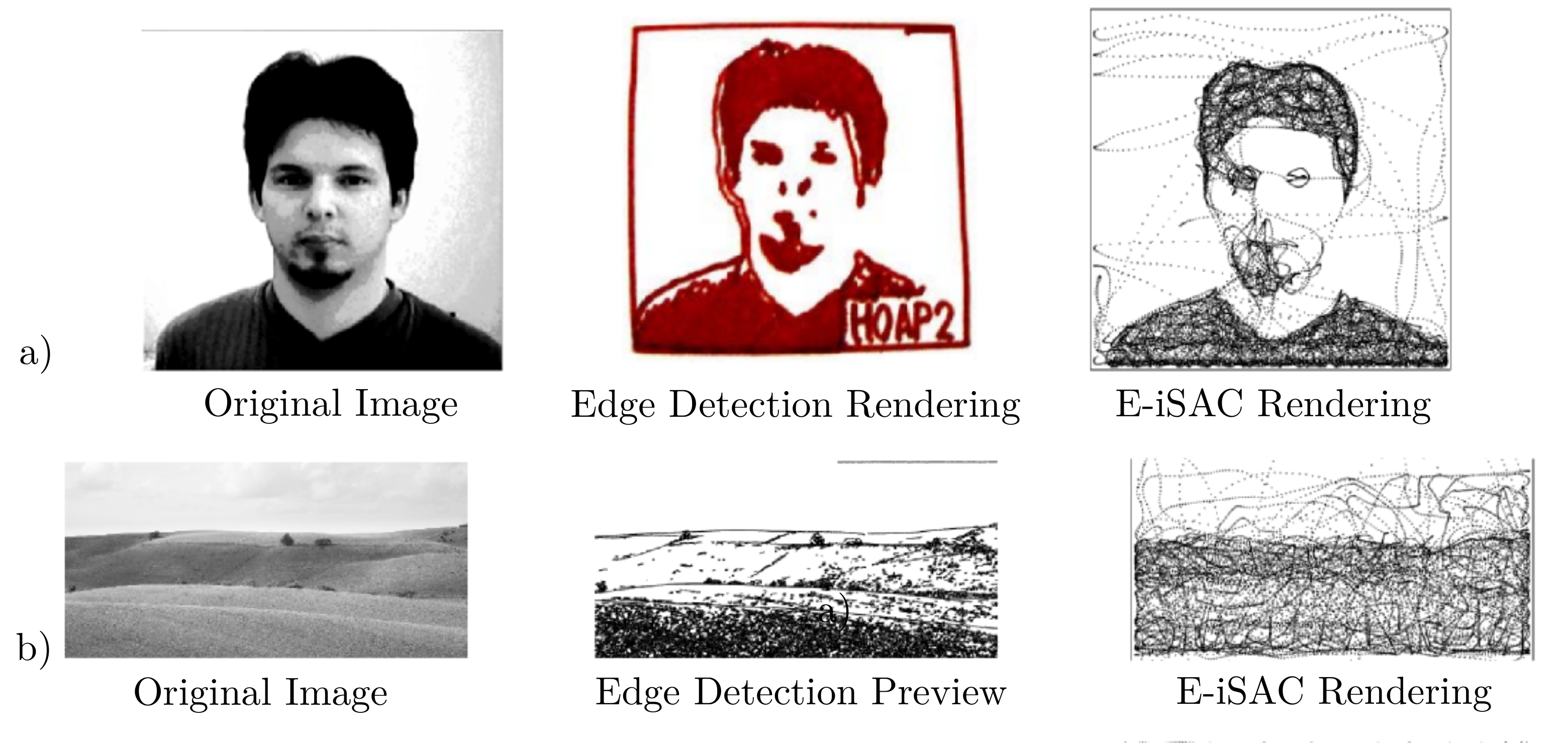}
    \caption{Comparison of the method in \cite{calinon2005humanoid} and the E-iSAC method. a) The original image and edge detection rendering using the HOAP2 robotic system come directly from \cite{calinon2005humanoid}. While the edge detection method is successful, producing a tractable trajectory requires morphological filters to extract the facial features and predetermined drawing primitives to render the shading.  b) Comparison for a landscape image, with a preview of the overlaid edge detection and binarization results compared to the E-iSAC trajectory. Producing a motion trajectory from the extracted contours of the preview would be computationally difficult (over 22000 contours result from the edge detection) or needs content-specific postprocessing,  and requires a precise drawing system. The E-iSAC rendering results in a highly tractable result. 
}\label{fig:EC_combined}
\end{figure} 

For comparison, we contrast ergodic control to the method employed in \cite{calinon2005humanoid},  a multi-stage process described in Figure \ref{fig:Method_chart}a, with a preliminary stage to render the outlines of the important features using edge detction and a secondary stage to render shading using image binarization. 
Figure \ref{fig:EC_combined}a shows the results of the method from \cite{calinon2005humanoid} compared to the trajectory created using the E-iSAC method for the double-order system. While the edge detection method from \cite{calinon2005humanoid} renders a successful recreation, obtaining a tractable trajectory requires parameter tuning and filtering to extract the most important features from the drawing, and predetermined drawing primitives and postprocessing to formulate the planar trajectory needed for rendering. Furthermore, the E-iSAC method is able to capture different levels of shading as opposed to the method in \cite{calinon2005humanoid} that only renders a binary image (black and white).

In addition, because the processing (e.g., filtering, parameter tuning) used by \cite{calinon2005humanoid} is tailored to drawing human portraits, the method is not robust to other content. To show this, we compare the results for rendering a landscape in Figure \ref{fig:EC_combined}b. While the simulated preview of the rendering appears successful, the image binarization to render shading fails as it is tuned specifically for human portraits. Instead, the quality of the image comes entirely from the edge detection step. However, processing and rendering the contours would be difficult (over 22,000 contours are generated to render the image), and the filtering implemented to make the results tractable are tailored specifically for facial portraits. While the E-iSAC method results in a more abstract image, the trajectory produced is tractable and the method is robust to a variety of subjects (as shown in Figure \ref{fig:imgcompspring}a). 

\begin{figure}[h!]
    \centering
    \includegraphics[width = \textwidth]{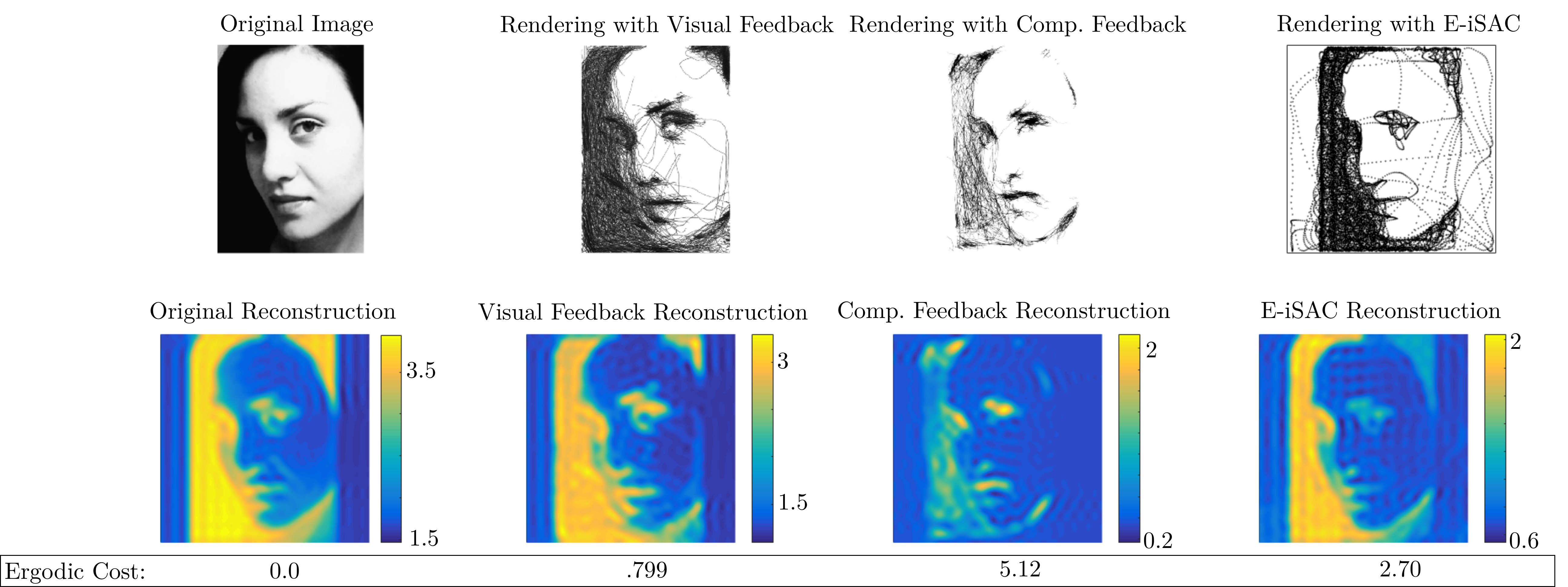}
    \caption{Comparison of drawings rendered with Paul the Robot \cite{tresset2013portrait} and the drawing rendered with E-iSAC and the ergodic Fourier reconstructions of these results. The original image and the images of the renderings from feedback come directly from \cite{tresset2013portrait}.  The E-iSAC rendering is able to perform comparably to the visual-feedback rendering (a closed-loop algorithm) and better than the computational-feedback rendering (an open-loop algorithm) with a simpler, open-loop algorithm. The reconstructions of the Fourier coefficients representing the different renderings with the respective ergodic costs show how ergodicity can be used as a quantitative metric for assessment of results. }\label{fig:Paul_comp}
\end{figure} 

Another drawing method is performed by Paul the robot \cite{tresset2013portrait}, which uses a complicated multi-stage process (shown in Figure \ref{fig:Method_chart}b) to render portraits. The first stage involves a salient-line extraction to draw the important features, and then performs a shading method using either visual feedback or computational feedback. The visual-feedback shading process uses an external camera to update the belief in real-time, while the computational-feedback shading process is based on the simulation of the line-extraction stage and is an open-loop process, similar to the E-iSAC method. Figure \ref{fig:Paul_comp} compares the results of Paul the robot \cite{tresset2013portrait} with the E-iSAC method, and shows the reconstructions of the Fourier coefficients representing each rendering. While the robot successfully renders the image, the E-iSAC method performs comparably well with a much simpler, open-loop algorithm and could be improved with the integration of an external camera setup to update the drawing in real-time. Furthermore, the method from \cite{tresset2013portrait} relies on a highly engineered system that moves precisely and cannot take into account a change in the robotic system, unlike E-iSAC. In addition, Figure \ref{fig:Paul_comp} shows the reconstructions of the Fourier representing each rendering, demonstrating how ergodicity can be used as a quantitative metric of assessment for the results, even if ergodic control is not used to determine the trajectories.    

\subsection{Baxter Experiments}
\label{sec:baxter_experiments}

We performed experiments executing the motion trajectories generated with the Baxter robot to demonstrate it physically rendering the portrait of Abraham Lincoln generated using the E-iSAC method from Section \ref{sec:Lincoln_example}. The Baxter robot is able to successfully complete the trajectory and render a recognizable portrait of Lincoln. The main stylistic differences between the simulation and the experimental results derive primarily from the assumption of an infinitesimally small marker point in the simulation and the board's inability to render shading. Improving the algorithm to enable encoding characteristics of the rendering tool (e.g., marker size, paintbrush stroke) and integration of an external camera to update the belief of current drawing would improve the rendering capabilities of the robotic system in the future. 

\begin{figure}[h!]
    \centering
    \includegraphics[width = .55\textwidth]{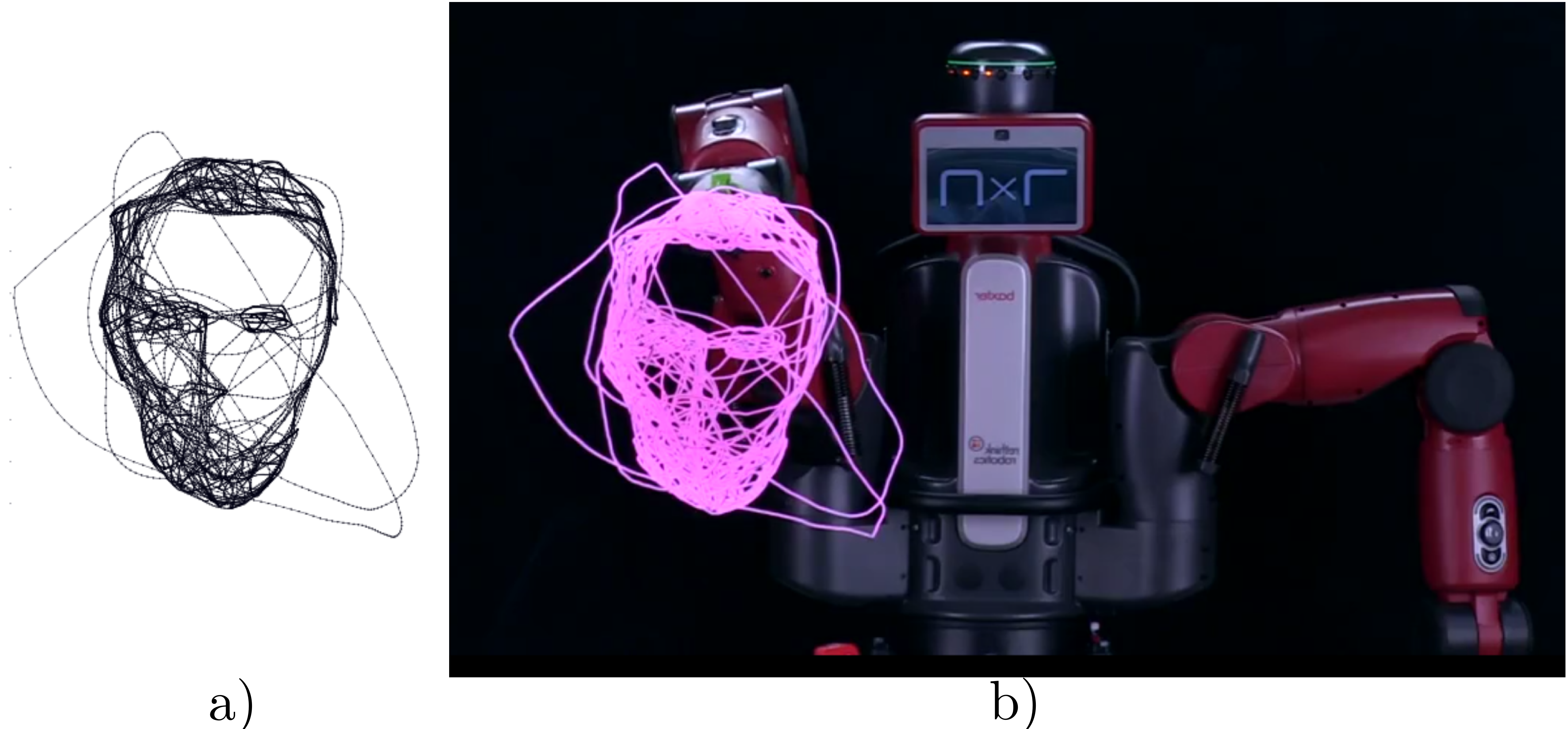}
    \caption{Experimental results with the Baxter robot. a) Trajectory generated by the E-iSAC method b) Rendering executed by the Baxter robot using the Lightboard \cite{lightboard}. The Baxter robot is successfully able to render the portrait of Abraham Lincoln using the motion trajectory generated. }\label{fig:Baxter_expandsim}
\end{figure} 

\section{Conclusion}
\label{sec:Conclusion}

This paper presented an autonomous process for translating visual information to physical motion. We demonstrate how ergodic metrics can be used as an actionable measure of symbolic spatial information, and explore the use of ergodicity as a measure that enables the robot to actively distinguish among different symbols with no prior knowledge of letter structure, other than the associated ergodic value.  In addition, ergodic control provides the robot the ability to naturally represent and recreate a wide range of visual inputs (from letters to portraits), while incorporating the robot's physical capabilities (e.g., dynamic drawing mechanics). Moreover, in the context of drawing, we see other drawing algorithms improve ergodicity, suggesting the use of ergodicity as a quantitative measure of assessment. Finally, we demonstrate experiments with the Baxter robot rendering these trajectories, and note that as optimal ergodic control can run in real-time, it can be ideal for developing interactive rendering behaviors in robots. In the future, we plan to adapt the algorithm to encode rendering characteristics of the system into the model and to integrate a visual feedback system to update the representation of the drawing in real-time. 

\section{Acknowledgements}
This material is based upon work supported by the National Science Foundation under grants CMMI 1334609 and IIS 1426961. Any opinions, findings, and conclusions or recommendations expressed in this material are those of the author(s) and do not necessarily reflect the views of the National Science Foundation.

\bibliographystyle{splncs03}
\bibliography{Literature_WAFR}

\end{document}
